\documentclass[letterpaper, 10 pt, conference]{ieeeconf}  

\IEEEoverridecommandlockouts                              

\usepackage{graphics} 
\usepackage{epsfig} 
\usepackage{mathptmx} 
\usepackage{times} 
\usepackage{amsmath} 
\usepackage{amssymb}  

\usepackage{multirow}
\usepackage{svg}

\usepackage{capt-of}
\usepackage{dsfont}

\usepackage{threeparttable}

\usepackage{algorithm}
\usepackage{algpseudocode}
\usepackage{wrapfig}
\usepackage{subcaption}
\usepackage{hyperref} 
\usepackage{cuted}
\usepackage{capt-of}
\usepackage{dsfont}

\def\titlePrefix{PlanarNeRF}

\title{\LARGE \bf
PlanarNeRF: Online Learning of Planar Primitives with Neural Radiance Fields
}

\author{Zheng Chen$^{1}$, Qingan Yan$^{2}$, Huangying Zhan$^{2}$, Changjiang Cai$^{2}$, Xiangyu Xu$^{2}$, Yuzhong Huang$^{3}$ \\ Weihan Wang$^{4}$, Ziyue Feng$^{5}$, Yi Xu$^{2}$ and Lantao Liu$^{1}$
\thanks{\newline
This work was partially done when Zheng Chen was an intern at OPPO US Research Center.
\newline
$^{1}$Z. Chen and L. Liu are with Luddy School of Informatics, Computing, and Engineering, Indiana University, Bloomington, IN 47408, USA. Email: {\tt\small \{zc11, lantao\}@iu.edu} 
\newline
$^{2}$Q. Yan, H. Zhan, C. Cai, X. Xu, and Y. Xu are with OPPO US Research Center.
\newline
$^{3}$Y. Huang is with University of Southern California.
\newline
$^{4}$W. Wang is with Stevens Institute of Technology.
\newline
$^{5}$Z. Feng is with Clemson University.
}
}

\begin{document}

\maketitle
\thispagestyle{empty}
\pagestyle{empty}

\begin{strip}\centering \vspace{-85pt}
{
\includegraphics[width=\linewidth]{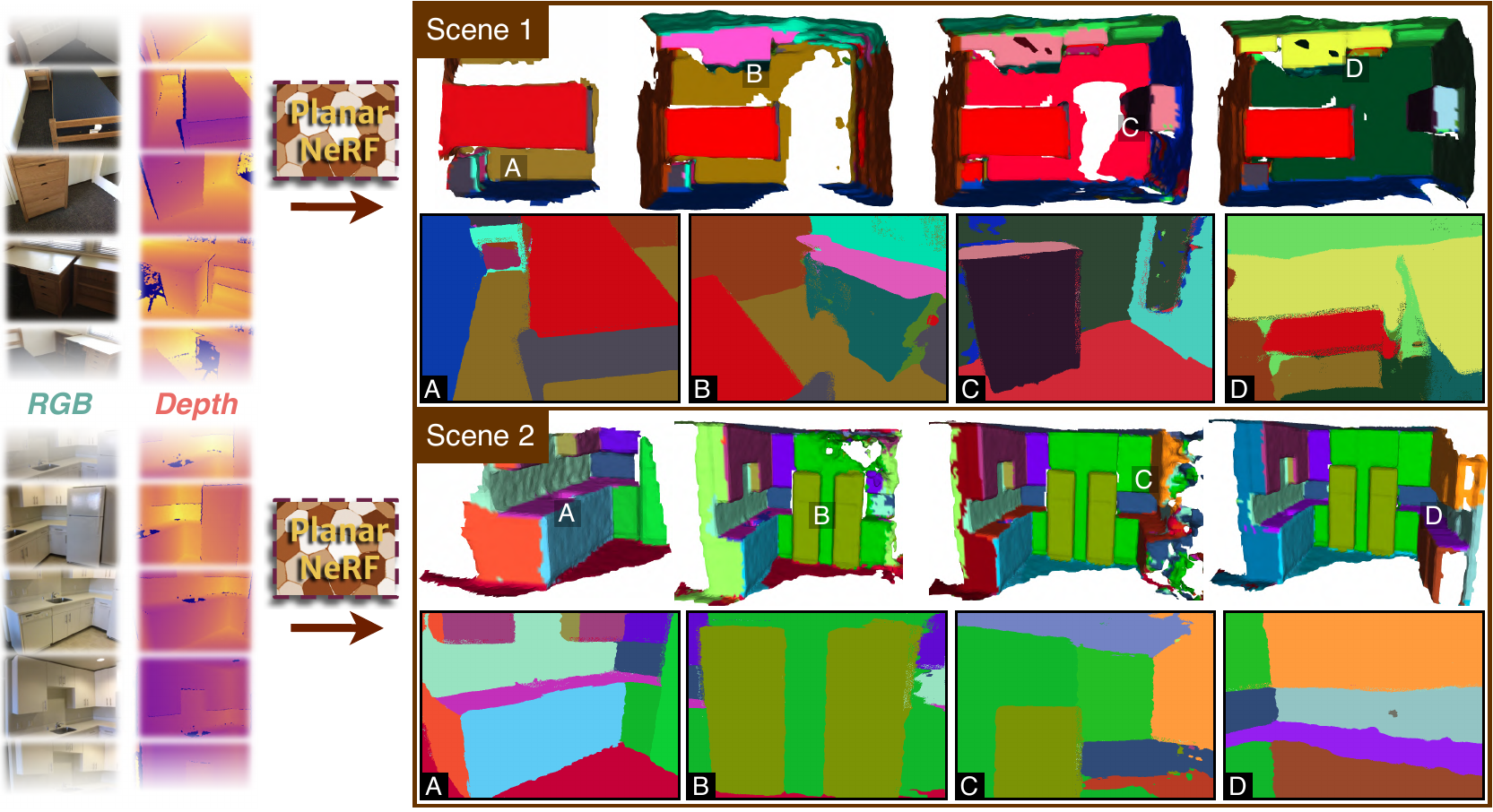} 
\captionof{figure}{
We introduce \textbf{\titlePrefix{}}, a framework designed to detect dense 3D planar primitives 
from monocular RGB and depth sequences. 
The method learns plane primitives in an online fashion while drawing knowledge from both scene appearance and geometry. 
Displayed are outcomes from two distinct scenes (Best viewed in color). 
Each case exhibits two rows: the top row visualizes the reconstruction progress, while the bottom row showcases \textbf{rendered} 2D segmentation images at different time steps. 
\vspace{-15pt}
\label{fig:title}}
}
\end{strip}

\begin{abstract}

Identifying spatially complete planar primitives from visual data is a crucial task in computer vision.
Prior methods are largely restricted to either 2D segment recovery or simplifying 3D structures, even with extensive plane annotations.
We present \titlePrefix{}, a novel framework capable of detecting dense 3D planes through online learning. 
Drawing upon the neural field representation, \titlePrefix{} brings three major contributions. 
First, it enhances 3D plane detection with concurrent appearance and geometry knowledge. 
Second, a lightweight plane fitting module is used to estimate plane parameters. 
Third, a novel global memory bank structure with an update mechanism is introduced, ensuring consistent cross-frame correspondence. 
The flexible architecture of \titlePrefix{} allows it to function in both 2D-supervised and self-supervised solutions, in each of which it can effectively learn from sparse training signals, significantly improving training efficiency.
Through extensive experiments, we demonstrate the effectiveness of \titlePrefix{} in various real-world scenarios and remarkable improvement in 3D plane detection over existing works.

\end{abstract}

\vspace{-10pt}
\section{Introduction}
\label{sec:intro}
Planar primitives stand out as critical elements in structured environments such as indoor rooms and urban buildings. 
Capturing these planes offers a concise and efficient representation,
and holds great impact across a spectrum of applications, including Virtual Reality, Augmented Reality, and robotic manipulation, etc. 
Beyond serving as a fundamental modeling block, planes are widely used in many data processing tasks, including object detection \cite{ren20213d}, registration \cite{li2021planar, zhu2016robust}, pose estimation \cite{chen2023monocular}, and SLAM \cite{zhou2023efficient, kaess2015simultaneous, kim2018linear}. 


Extensive efforts have been dedicated to exploring different plane detection methodologies. 
Nevertheless, notable limitations persist within current approaches. 
First, many of them produce only isolated per-view 2D plane segments~\cite{liu2019planercnn, yu2019single, tan2021planetr}. 
Although certain methods~\cite{jin2021planar,agarwala2022planeformers,tan2023nope} establish correspondences across sparse (typically two) views, they still lack spatial consistency, leading to incomplete scene representations. 
Recently, an end-to-end deep model \cite{xie2022planarrecon} was introduced for 3D plane detection; however, its outcomes tend to oversimplify scene structures. 
Moreover, the aforementioned models heavily rely on extensive annotations --- pose, 2D planes, and 3D planes --- consequently limiting their generalization capabilities. 
While fitting-based methods like \cite{rabbani2006segmentation,schnabel2007efficient} operate without annotations, they are typically restricted to offline detection, involving heavy iterations and posing computational challenges.  

We propose \titlePrefix{}, an online 3D plane detection framework (Fig.~\ref{fig:title}) that overcomes the above limitations. 
Specifically, we extend the neural field representation to regress plane primitives with both appearance and geometry for more complete and accurate results.  
The framework's efficient network design allows for dual operational modes:
\textbf{\titlePrefix{}-S}, a supervised mode leveraging sparse 2D plane annotations;
and \textbf{\titlePrefix{}-SS}, a self-supervised mode that extracts planes directly from depth images. 
In \titlePrefix{}-SS, we adopt RANSAC~\cite{schnabel2007efficient} for estimating \textit{local} plane instances over
a highly sparse set of sampled points in each iteration, leading to a \textit{lightweight} plane fitting module.
Then a global memory bank is maintained to ensure consistent tracking of plane instances across different views and to generate labels for the sparse points.
The inherent multi-view consistency and smoothness of NeRF facilitate the propagation of sparse labels. 

\vspace{-5pt}
\section{Related Work}
\label{sec:related}
\noindent{\textbf{Single View Plane Detection.}} 
Many studies focus on directly segmenting planes from individual 2D images. PlaneNet~\cite{liu2018planenet} was among the first to encapsulate the detection process within an end-to-end framework directly from a single image. Conversely, PlaneRecover~\cite{yang2018recovering} introduces an unsupervised approach for training plane detection networks using RGBD data.
Meanwhile, PlaneRCNN~\cite{liu2019planercnn} capitalizes on Mask-RCNN's~\cite{he2017mask} generalization capability to identify planar segmentation in input images, simultaneously regressing 3D plane normals from fixed normal anchors. In contrast, PlaneAE~\cite{yu2019single} assigns each pixel to an embedding feature space, subsequently grouping similar features through mean-shift algorithms. Additionally, PlaneTR~\cite{tan2021planetr} harnesses line segment constraints and employs Transformer decoders~\cite{carion2020end} to enhance performance further. Despite these advancements, 
the detected plane instances lack consistency across different frames.

\noindent{\textbf{Multi-view Plane Detection.}} SparsePlanes~\cite{jin2021planar} detects plane segments in two views and uses a deep neural network architecture with an energy function for correspondence optimization.
PlaneFormers~\cite{agarwala2022planeformers}, eschewing handcrafted energy optimization, introduces a Transformer architecture to directly predict plane correspondences. 
NOPE-SAC~\cite{tan2023nope} associates two-view camera pose estimation with plane correspondence in the RANSAC paradigm while enabling end-to-end learning. 
PlaneMVS~\cite{liu2022planemvs} unifies plane detection and plane MVS with known poses and facilitates mutual benefits between these two branches.
Although multi-view inputs enhance segmentation consistency, they still lack a global association, preventing the construction of complete scenarios. 
PlanarRecon~\cite{xie2022planarrecon} progressively fuses multi-view features and extracts 3D plane geometries from monocular videos in an end-to-end fashion, bypassing per-view segmentation. 
Nonetheless, it necessitates 3D ground truth prerequisites and tends to oversimplify the resulting output.


\noindent{\textbf{Neural Scene Reconstruction.}} The groundbreaking NeRF~\cite{mildenhall2021nerf} introduced an innovative solution for 3D environment representation, upon which numerous studies have demonstrated outstanding performance in scene reconstruction~\cite{wang2021neus, wang2023neus2, wang2022neuralroom, azinovic2022neural, yu2022monosdf, wu2022voxurf, stier2023finerecon, li2023neuralangelo, ye2023self, gao2023surfelnerf}. In particular, Nice-SLAM~\cite{zhu2022nice} builds a series of learnable grid architectures serving as hierarchical feature encoders and conducts pose optimization and dense mapping. Nicer-SLAM~\cite{zhu2023nicer} refines this approach by reducing the necessity for depth images and achieves comparable reconstruction results. Co-SLAM~\cite{wang2023co} adopts hash maps instead of grids as the feature container and introduces coordinate and parametric encoding for expedited convergence and querying.

\begin{figure*}[t]
  \centering
   \includegraphics[width=0.87\linewidth]{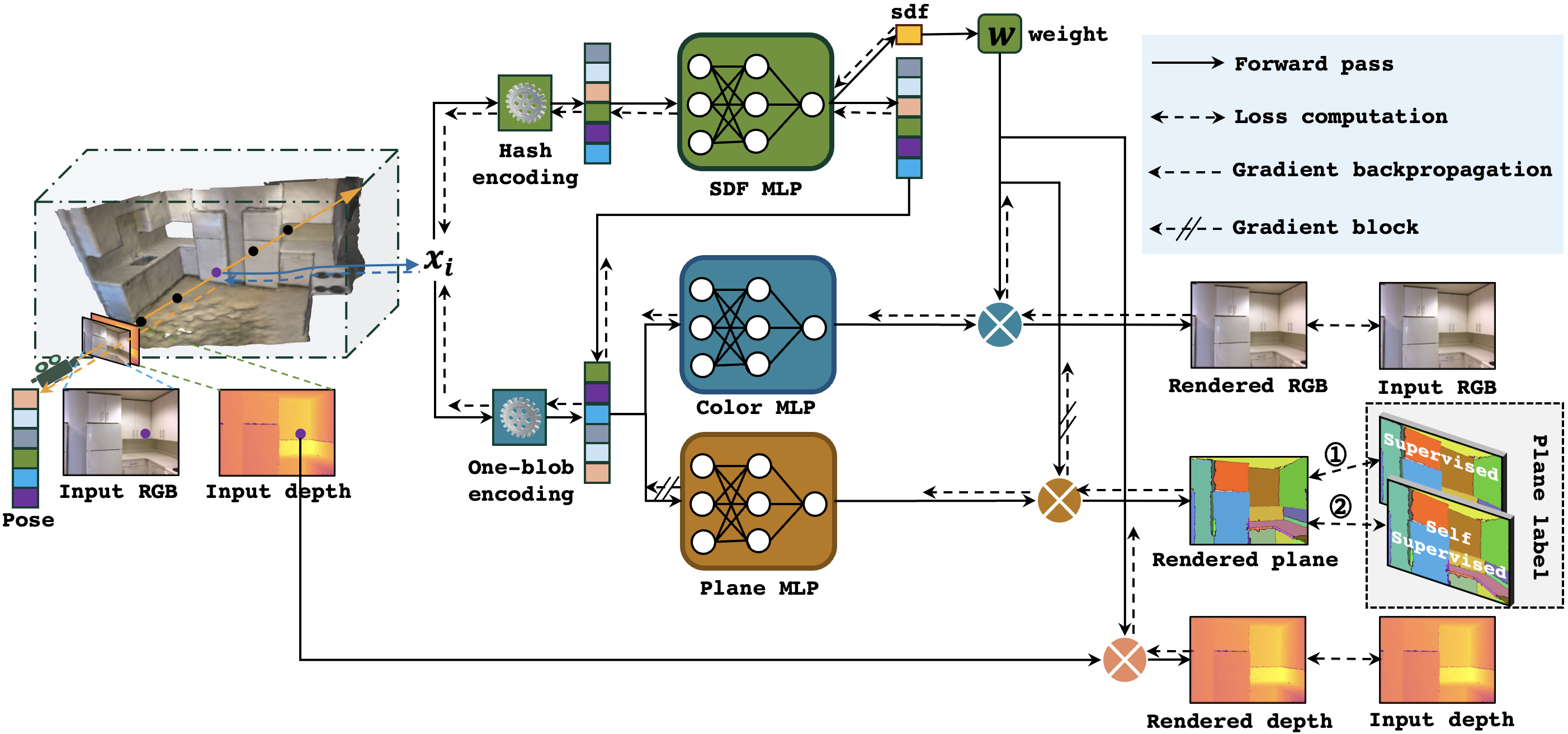} 
   \caption{\small  
   \textbf{Overview of \titlePrefix{}}. 
   \titlePrefix{} processes monocular RGB and depth image sequences, enabling online pose estimation. 
   It offers two modes: \textcircled{\scriptsize{1}} \titlePrefix{}-S (supervised) with 2D plane annotations, and 
   \textcircled{\scriptsize{2}} \titlePrefix{}-SS (self-supervised) without annotations. 
   The framework includes an efficient plane fitting module and a global memory bank for consistent plane labeling.
   \vspace{-18pt}
   }
   \label{fig:overall_framework}
\end{figure*}

\vspace{-3pt}
\section{Methodology}
\vspace{-5pt}
\label{sec:method}
\subsection{Preliminaries}
NeRF (Neural Radiance Fields)~\cite{mildenhall2021nerf} conceptualizes a scene as a continuous function, typically represented by a multi-layer perceptron (MLP).
This function, defined as 
$F(\textbf{x}, \textbf{v}) \mapsto (\textbf{c}, \sigma)$,
maps a 3D point $\textbf{x}$ and a 2D viewing direction \( \mathbf{v} \) to the corresponding RGB color \( \mathbf{c} \) and volume density \( \sigma \).
For a ray \( R(t) = \mathbf{o} + t\mathbf{v} \) with origin \( \mathbf{o} \), the rendered color \( \mathbf{C}(R) \) is obtained by integrating points along the ray via volume rendering:
\begin{equation}
    \label{eq:original_color_rendering}
    \mathbf{C}(R) = \int_{t_n}^{t_f} T(t) \sigma(R(t)) \mathbf{c}(R(t), \mathbf{v}) \, dt,
\end{equation}
where \( T(t) = \exp\left(-\int_{t_n}^{t} \sigma(R(s)) \, ds\right) \) 
is the accumulated transmittance from the near bound \( t_n \) to \( t \), and \( t_f \) is the far bound. 

Recent advancements in NeRF have enhanced rendering and reconstruction by modifying the original framework. 
Key improvements include 
using Signed Distance Fields (SDF) for predictions \cite{wang2021neus}, 
employing separate neural networks for RGB and geometry with augmented inputs \cite{SunSC22}, 
recalculating weights in the rendering equation based on SDF \cite{azinovic2022neural}, 
and adopting hash and one-blob encoding for positional data \cite{muller2022instant}. 
Additionally, depth rendering is used to improve geometry learning \cite{yu2022monosdf}. 
In \titlePrefix{}, we incorporate these recent modifications, resulting in an updated and optimized color rendering equation:
\begin{equation}
    \label{eq:new_color_rendering}
    \mathbf{C}(R) = \frac{1}{\sum_{i=1}^{M} w_i} \sum_{i=1}^{M} w_i \mathbf{c}_i(R(t), \mathbf{v}),
\end{equation}
where $M$ is the number of sampled points along the ray, 
and $w_i$ is the weight computed based on SDF: $w_i = \sigma\left(\frac{s_i}{tr}\right) \sigma\left(-\frac{s_i}{tr}\right)$.
Here, $s_i$ is the predicted SDF values along the ray; 
$tr$ is a predefined truncation threshold for SDF; 
and $\sigma(\cdot)$ is the sigmoid function. 
Similar to \ref{eq:new_color_rendering}, 
the rendering equation for depth is:
\begin{equation}
    \label{eq:depth_rendering}
    D(R) = \frac{1}{\sum_{i=1}^{M} w_i} \sum_{i=1}^{M} w_i dp_i(R(t), \mathbf{v}),
\end{equation}
where $dp_i$ is the depth of sampled points along the ray.

\vspace{-2pt}
\subsection{Framework Overview}
\vspace{-2pt}
The overview of \titlePrefix{} is depicted in Fig. \ref{fig:overall_framework}. 
Alongside SDF and color rendering branches, an additional plane rendering branch (Section \ref{sec:method:plane_rendering}) is introduced to map 3D coordinates to 2D plane instances, utilizing appearance and geometry prior. 
The plane MLP and color MLP share the same input, which combines a one-blob encoded 3D coordinate and a learned SDF feature vector. 
In \titlePrefix{}-S, while consistent 2D plane annotations are requisite, they are often unavailable in real-world scenarios, where manual labeling for plane instance segmentation is costly. 
To tackle this challenge, we use RANSAC~\cite{schnabel2007efficient} to estimate plane parameters from depth images and propose a global memory bank (Section \ref{sec:bank}) to track consistent planes and produce plane labels. 
During the training phase, gradient backpropagation from the plane branch to the SDF is blocked to prevent potential negative impacts on geometry learning, with further qualitative analysis provided in Section \ref{sect:ablation}.

\subsection{Plane Rendering Learning}
\label{sec:method:plane_rendering}
Similar to  Eq. (\ref{eq:new_color_rendering}) and Eq. (\ref{eq:depth_rendering}), we propose the rendering equation for planes as:
\begin{equation}
    \label{eq:plane_rendering}
    \mathbf{P}(R) = \frac{1}{\sum_{i=1}^{M} w_i} \sum_{i=1}^{M} w_i \mathbf{p}_i(R(t), \mathbf{v}),
\end{equation}
where $\mathbf{p}_i$ is the plane classification probability vector of sampled points along the ray. 

Conventionally, instance segmentation learning has been approached using either anchor boxes \cite{he2017mask, liu2019planercnn} or a bipartite matching \cite{tan2021planetr, cheng2021per, cheng2022masked, siddiqui2023panoptic}.
Anchor boxes-based methods often involve complex pipelines with heuristic designs. 
In contrast, bipartite matching-based methods establish an optimized correspondence between predictions and ground truths before computing the loss. 
The instance segmentation loss based on bipartite matching can be expressed as: 
\begin{equation}
    \label{eq:ins_loss}
    \mathcal{L}_{ins} = -\frac{1}{Q}\sum_{q=1}^Q \sum_{c=1}^C y_c \log \hat{y}_c,
\end{equation}
where $Q$ is the number of pixels; $C$ is the number of classes. $y_c$ is the c$^{th}$ element in the ground truth label $\mathbf{y}$, and $y_c = \mathds{1}_{\{c=m(\hat{\mathbf{y}}, \mathbf{y})\}}$, where $m(\cdot)$ is the matching function, and the assignment cost can be given by the intersection over union of each instance between the prediction and the ground truth. $\hat{y}_c$ is the c$^{th}$ element in the prediction probability vector $\hat{\mathbf{y}}$. Using bipartite matching stems from the inherent discrepancies in index values between instance segmentation predictions and the ground truth labels. 
We only need to match the segmented area and distinguish one instance from another.

In contrast to the instance segmentation methods previously discussed, \titlePrefix{} employs a distinct approach for plane instance segmentation. We adopt a \textbf{fixed matching} technique, akin to that used in semantic segmentation, to compute the segmentation loss. This method is chosen because our primary objective is to learn consistent 3D plane instances. Consequently, it is imperative that the rendered 2D plane instance segmentation remains consistent across different frames. To uphold this consistency, we ensure that the indices in the predictions strictly match the values provided in the ground truth during loss computation.

\subsection{Global Memory Bank}
\label{sec:bank}
We use RANSAC~\cite{schnabel2007efficient} for estimating local plane instances.
Plane estimations in different iterations are independent of each other. This lacks consistency as new data constantly comes in. We propose a novel global memory bank to maintain the plane parameters across different frames. 



The key part of maintaining the bank is the similarity measure between two planes. Based on RANSAC, we are able to obtain the plane vector for each plane instance. An intuitive way to compare two vectors is to compute the Euclidean distance, $\left\| \mathbf{p}_1 - \mathbf{p}_{2} \right\|_2$. However, this way fails for planes because each element in the plane parameter vector has a physical meaning. A reasonable way to compare the distance between two plane parameters is:
\begin{equation}
    \label{eq:plane_dist_1}
    dist^{'}(\mathbf{p}_1, \mathbf{p}_2) = 1 - \left| \frac{\left< \mathbf{n}_1, \mathbf{n}_2 \right>}{\left\| \mathbf{n}_1\right\|\left\|\mathbf{n}_2 \right\|} \right| + \left| d_1 - d_2 \right|,~~d_1, d_2 \in \mathbb{R}_{\geq 0},
\end{equation}
where $\mathbf{p}_1 = \left [ \mathbf{n}_1, d_1 \right ]^T, \mathbf{p}_2 = \left [ \mathbf{n}_2, d_2 \right ]^T$. All offset values must be non-negative because a plane parameter vector and its negative version describe the  same plane spatially, ignoring the normal orientations.

Unfortunately, Eq. (\ref{eq:plane_dist_1}) works well as a similarity measure but it is too sensitive to the estimation noises.
Directly comparing two plane vectors lacks the robustness to the noises. To tackle this issue, we propose to use a simple yet robust way to compute the similarity measure. There are two representations for one plane --- the plane parameters ($\mathbf{p}_i$) or the points ($PO_i = \{ \mathbf{po}_j \}_{j=0}^{n_j}$) belonging to the plane instance. The new similarity measure is based on the distance between \textbf{points to the plane}. Assume we use $\mathbf{p}_1$ to represent one plane and $PO_2$ for another, then we can have:
\begin{equation}
    \label{eq:plane_dist_2}
    dist(\mathbf{p}_1, \mathbf{p}_2) = \frac{1}{n_j} \sum_{j=1}^{n_j} \frac{\left| n_1^x\cdot x_2 + n_1^y\cdot y_2+ n_1^z\cdot z_2 - d_1 \right|}{\left ( (n_1^x)^2 + (n_1^y)^2  + (n_1^z)^2\right )^{\frac{1}{2}}}.
\end{equation}

If a new plane is found highly similar to one of the plane vectors inside the bank, i.e., $dist(\mathbf{p}_{new}, \mathbf{p}_{bank})<\tau_{dist}$, where $\tau_{dist}$ is the distance threshold for decisions, then we return the index of $\mathbf{p}_{bank}$ in the bank as the index annotation for the sampled points belonging to the plane instance $\mathbf{p}_{new}$. Otherwise, we add the $\mathbf{p}_{new}$ into the bank. The plane label is given by $y_c = \mathds{1}_{\{c=k\}}$ (see Eq.~(\ref{eq:ins_loss})). If \titlePrefix{}-S is used, then $y_c$ is assumed to be known.

To further increase the robustness of the global memory bank, we use the Exponential Moving Average (EMA) to update the plane parameters stored in the bank if the highly similar plane in the bank is found:
\begin{equation}
    \label{eq:ema}
    \mathbf{p}_{bank} = \psi \mathbf{p}_{new} + (1-\psi)\mathbf{p}_{bank},
\end{equation}
where $\psi$ is the EMA coefficient. Note that before the update using EMA, the offset values must satisfy the constraint in Eq. (\ref{eq:plane_dist_1}). 

\section{Experiments}
\label{sec:exp}

\begin{figure} [t]
{
  \centering
   \includegraphics[width=\linewidth]{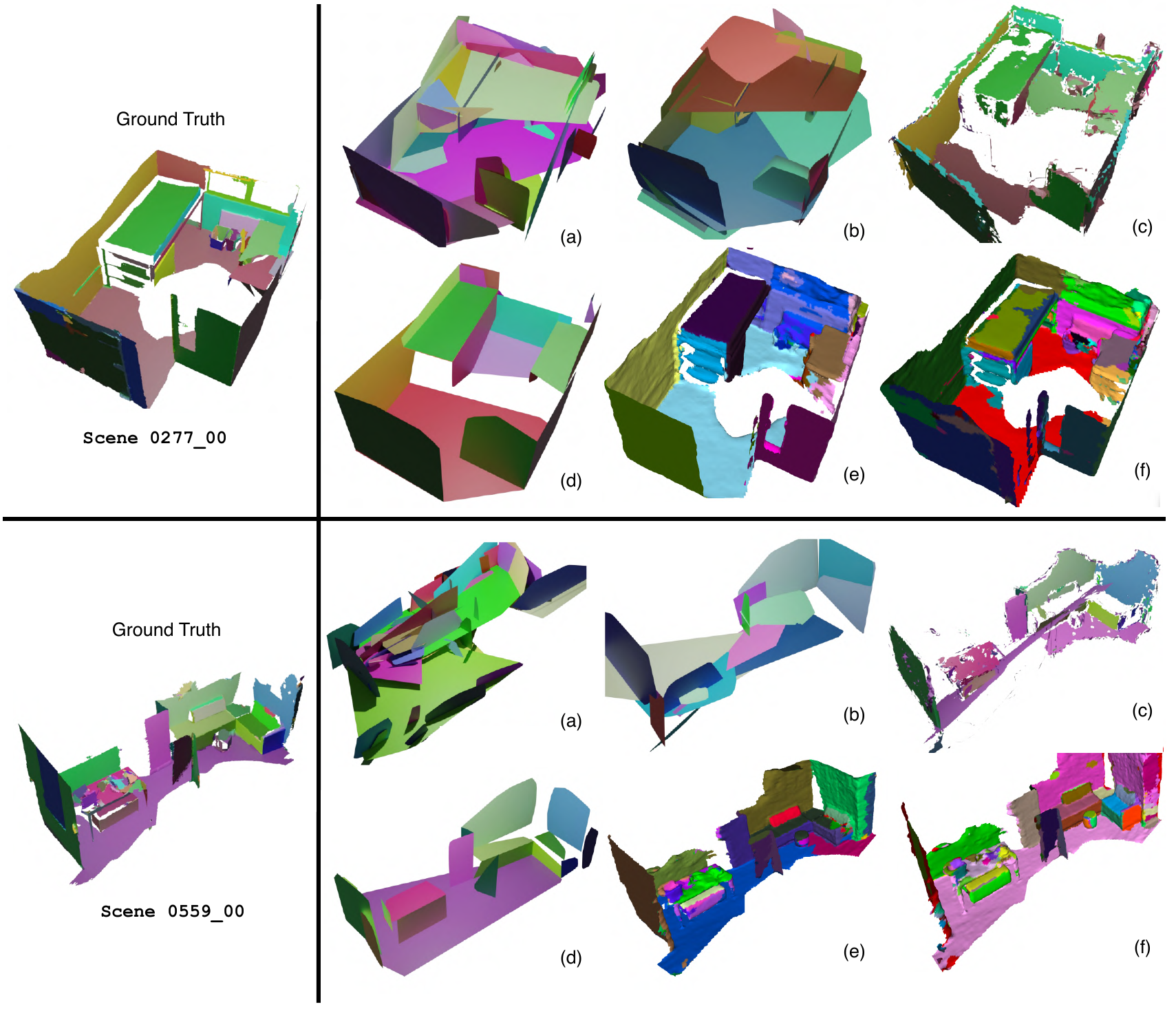} 
   \caption{\small Qualitative comparisons of different methods for two scenes. (a) PlaneAE; (b) ESTDepth+PEAC; (c) NeuralRecon+Seq-RANSAC; (d) PlanarRecon; (e) PlanarNeRF-SS (ours); and (f) PlanarNeRF-S (ours). \vspace{-18pt}
   } 
   \label{fig:qual_1}
   }
\end{figure}

\begin{table*}[t]
\centering
\resizebox{\textwidth}{!}{%
\begin{threeparttable}
\caption{Comparisons for 3D geometry, memory and speed on ScanNet. \textbf{\color{purple}Red} for the best and \textbf{\color{teal}green} for the second best (same for the following). 
$\nmid$:
Since our method is an online learning method, we report the \textit{memory} and \textit{time} used during \textbf{training}. Others are offline-trained, hence \textbf{inference}. 
$\star$:
For \textbf{\titlePrefix{}-S}. \textbf{SS} and \textbf{S} share the same geometry learning and GPU memory. The time gap is mainly caused by the self-supervised plane label generation (on CPU) in \textbf{SS}.  
}
\label{tab:3d_geo}
\begin{tabular}{l|c|ccccc|c|c}
\hline \hline 
 Method & Val. Set & Acc.~$\downarrow$ & Comp.~$\downarrow$ & Recall~$\uparrow$ & Prec.~$\uparrow$ & F-score~$\uparrow$ & Mem. (GB)~$\downarrow$ & Time (ms)~$\downarrow$ \\
\hline
NeuralRecon~\cite{sun2021neuralrecon} + Seq-RANSAC~\cite{fischler1981random} & \multirow{5}{*}{\cite{murez2020atlas}} & 0.144 & 0.128 & 0.296 & 0.306 & 0.296 & 4.39 & 586 \\
Atlas~\cite{murez2020atlas} + Seq-RANSAC~\cite{fischler1981random} &  & 0.102 & 0.190 & 0.316 & 0.348 & 0.331 & 25.91 & 848 \\
ESTDepth~\cite{long2021multi} + PEAC~\cite{feng2014fast} &  & 0.174 & 0.135 & 0.289 & 0.335 & 0.304 & 5.44 & \textbf{\color{teal}101} \\
PlanarRecon~\cite{xie2022planarrecon} & & \textbf{\color{teal}0.154} & \textbf{\color{teal}0.105} & \textbf{\color{teal}0.355} & \textbf{\color{teal}0.398} & \textbf{\color{teal}0.372} & \textbf{\color{teal}4.43} & \textbf{\color{purple}40}\\
\textbf{\titlePrefix{}-SS (Ours)} & & \textbf{\color{purple}0.059} & \textbf{\color{purple}0.073} & \textbf{\color{purple}0.661} & \textbf{\color{purple}0.651} & \textbf{\color{purple}0.654} & \textbf{\color{purple}4.09}\tnote{$\nmid$} & 328\tnote{$\nmid$}~~/~131\tnote{$\nmid$}~\tnote{$\star$}\\
\hline 
PlaneAE~\cite{yu2019single} & \multirow{3}{*}{\cite{yu2019single}} & \textbf{\color{teal}0.128} & 0.151 & 0.330 & 0.262 & 0.290 & 6.29 & \textbf{\color{purple}32} \\
PlanarRecon~\cite{xie2022planarrecon} & & 0.143 & \textbf{\color{teal}0.098} & \textbf{\color{teal}0.372} & \textbf{\color{teal}0.412} & \textbf{\color{teal}0.389} & \textbf{\color{teal}4.43} & \textbf{\color{teal}40} \\
\textbf{\titlePrefix{}-SS (Ours)} & & \textbf{\color{purple}0.063} & \textbf{\color{purple}0.078} & \textbf{\color{purple}0.674} & \textbf{\color{purple}0.657} & \textbf{\color{purple}0.665} & \textbf{\color{purple} 4.09}\tnote{$\nmid$} & 328\tnote{$\nmid$}~~/~131\tnote{$\nmid$}~\tnote{$\star$}\\
\hline \hline
\end{tabular} 
\end{threeparttable}
}
\end{table*}

\subsection{Baselines and Evaluation Metrics}
\titlePrefix{} has two working modes: \textbf{\titlePrefix{}-S} where 2D plane annotations are used; and \textbf{\titlePrefix{}-SS} where no annotations are used. We compare our method with four types of approaches: (1) Single view plane recovering \cite{yu2019single}; (2) Multi-view depth estimation \cite{long2021multi} with depth based plane detection \cite{feng2014fast}; (3) Volume-based 3D reconstruction \cite{sun2021neuralrecon} with Sequential RANSAC \cite{fischler1981random}; and (4) Learning-based 3D planar reconstruction \cite{xie2022planarrecon}.

\begin{table}
\centering
\caption{3D plane instance segmentation comparison on ScanNet. } 
\label{tab:3d_ins}
\resizebox{\columnwidth}{!}{ %
\begin{tabular}{l|ccc}
\hline \hline 
 Method & VOI~$\downarrow$ & RI~$\uparrow$ & SC~$\uparrow$\\
\hline
NeuralRecon~\cite{sun2021neuralrecon} + Seq-RANSAC~\cite{fischler1981random} & 8.087 & 0.828 & 0.066  \\
Atlas~\cite{murez2020atlas} + Seq-RANSAC~\cite{fischler1981random} & 8.485 & 0.838 & 0.057 \\
ESTDepth~\cite{long2021multi} + PEAC~\cite{feng2014fast} & 4.470 & 0.877 & 0.163 \\
PlanarRecon~\cite{xie2022planarrecon} & 3.622 & 0.897 & \textbf{\color{teal}0.248} \\
\textbf{\titlePrefix{}-SS (Ours)} & \textbf{\color{teal}2.940} & \textbf{\color{teal}0.922} & 0.237 \\
\textbf{\titlePrefix{}-S (Ours)} & \textbf{\color{purple}2.737} & \textbf{\color{purple}0.937} & \textbf{\color{purple}0.251} \\
\hline 
PlaneAE~\cite{yu2019single} & 4.103 & 0.908 & 0.188\\
PlanarRecon~\cite{xie2022planarrecon} & 3.622 & 0.898 & \textbf{\color{teal}0.247}\\
\textbf{\titlePrefix{}-SS (Ours)} & \textbf{\color{teal}2.952} & \textbf{\color{teal}0.928} & 0.235 \\
\textbf{\titlePrefix{}-S (Ours)} & \textbf{\color{purple}2.731} & \textbf{\color{purple}0.940} & \textbf{\color{purple}0.252}\\
\hline \hline
\end{tabular} 
} 
\end{table}

Following the baseline work \cite{xie2022planarrecon}, we evaluate the performance of our method in terms of both geometry as well as plane instance segmentation. More specifically, for geometry evaluation, we use five metrics \cite{murez2020atlas}: Completeness; Accuracy; Recall; Precision; and F-score. For plane instance segmentation, we use three metrics \cite{liu2019planercnn}: Rand Index (RI); Variation of Information (VOI); and Segmentation Covering (SC).

\subsection{Datasets and Implementations}
Our experiments are conducted using the ScanNetv2 dataset \cite{dai2017scannet}. This dataset is comprised of RGB-D video sequences captured with a mobile device across 1,613 different indoor scenes. Due to the lack of ground truth data in the test set, following the previous work \cite{xie2022planarrecon}, we adopt the approach used by PlaneRCNN \cite{liu2019planercnn}, creating 3D plane labels for both training and validation datasets. To be consistent with the previous work, we also assess our method's performance on two distinct validation sets, which are differentiated by the scene splits previously employed in works \cite{yu2019single, murez2020atlas}.

\begin{figure}[t]
  \centering
   \includegraphics[width=\linewidth]{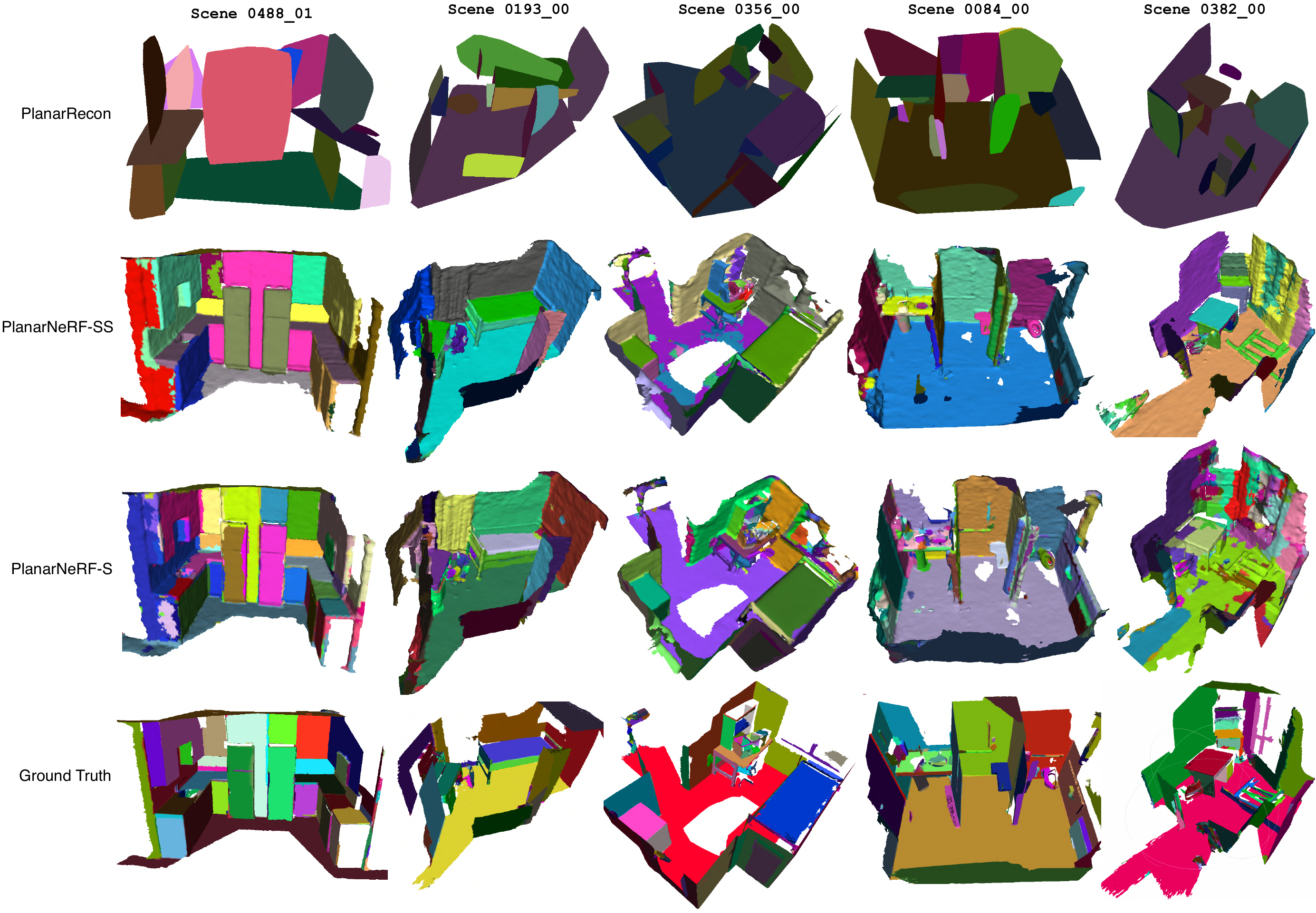} 
   \caption{\small Qualitative comparison between the recent SOTA --- PlanarRecon \cite{xie2022planarrecon} and ours on ScanNet.
   \vspace{-15pt}
   }
   \label{fig:qual_2}
\end{figure}

Besides ScanNetv2, we test our method on two additional datasets: Replica \cite{straub2019replica} and Synthetic scenes from NeuralRGBD \cite{azinovic2022neural}. 
As baselines lack reported results on these datasets, we present only our model's outcomes. 
Detailed information about these datasets is available in the supplementary material. 
We employ Co-SLAM \cite{wang2023co} as our backbone, with further implementation details provided in the supplementary material.

\vspace{-5pt}
\subsection{Qualitative Results}
We show \textbf{qualitative} comparisons between our method and all the baselines in Fig. \ref{fig:qual_1}, where the results of two scenes in ScanNet are presented. Different color represents different plane instance. Note that the colors in predictions do not necessarily match the ones in the ground truths. PlaneAE is able to reconstruct the single-view planes but fails to organize them in 3D space consistently. ESTDepth + PEAC is better than PlaneAE but still suffers from a lack of consistency. NeuralRecon + Seq-RANSAC can produce good plane estimations but the geometry is poor and therefore diminishes the performance of instance segmentation. PlanarRecon can generate consistent and compact 3D planes but the results are oversimplified and many details of the rooms are missed. We can easily see that the results of our method are significantly superior to others in terms of both geometry and instance segmentation. \titlePrefix{}-S can generate plane instance segmentation \textbf{highly close} to the ground truth when only 2D plane annotations are used. \titlePrefix{}-SS also shows a high-standard segmentation quality even though no any annotations are used. If we consider a comparison in the space of plane parameters, i.e., planes sharing highly similar parameters are classified as one plane instance, our \titlePrefix{}-SS gains more credits.

\begin{table*}[t] 
\centering
\caption{Ablation studies for similarity threshold and EMA coefficient.}  
\label{tab:ablation_studies} 
\begin{minipage}{0.48\linewidth}
\centering
\subcaption{Similarity threshold} 
\label{tab:similarity_threshold}
\begin{tabular}{lcccccc}
\hline \hline 
 $\tau_{dist}$ & 0.01 & 0.1 & 0.2 & 0.3 & 0.5 & 0.7 \\
\hline
VOI~$\downarrow$ & 3.219 & \textbf{\color{purple}2.726} & 2.951 & \textbf{\color{teal}2.753} & 3.356 & 3.244\\
\hline
RI~$\uparrow$ & \textbf{\color{teal}0.878} & 0.874 & 0.875 & \textbf{\color{purple}0.880} & 0.858 & 0.856 \\
\hline
SC~$\uparrow$ & 0.251 & \textbf{\color{purple}0.338} & 0.276 & \textbf{\color{teal}0.279} & 0.200 & 0.141 \\
\hline \hline
\end{tabular}
\end{minipage}
\hfill
\begin{minipage}{0.48\linewidth}
\centering
\subcaption{EMA coefficient} 
\label{tab:ema}
\begin{tabular}{lcccccc}
\hline \hline 
 $\psi$ & 0.6 & 0.7 & 0.8 & 0.9 & 0.99 & 0.999 \\
\hline
VOI~$\downarrow$ & 3.532 & 3.655 & 3.587 & \textbf{\color{teal}3.018} & 3.438 & \textbf{\color{purple}2.812}\\
\hline
RI~$\uparrow$ & 0.830 & 0.814 & 0.879 & \textbf{\color{teal}0.881} & \textbf{\color{purple}0.890} & 0.866\\
\hline
SC~$\uparrow$ & 0.204 & 0.162 & 0.088 & 0.146 & \textbf{\color{teal}0.268} & \textbf{\color{purple}0.314}\\
\hline \hline 
\end{tabular}
\end{minipage} 
\end{table*} 

We also present \textbf{quantitative} comparisons for geometry quality (Table \ref{tab:3d_geo}) and instance segmentation (Table \ref{tab:3d_ins}). From Table \ref{tab:3d_geo}, we can see that our method achieves systematic superiority to others in all geometry metrics with very low GPU memory consumption. \titlePrefix{} is not as fast as PlanarRecon and ESTDepth+PEAC because our method is an online-learning method; The training of SDF and color rendering takes around $180ms$ while self-supervised plane estimation and plane rendering learning takes around $148ms$. It is acceptable to be slower than the pure inference speed of the offline-trained models. From Table \ref{tab:3d_ins}, we can still see the advantages of our method over other baselines in terms of the quality of plane instance segmentation.
\begin{figure}[t]
  \centering
  \begin{subfigure}[b]{0.8\linewidth}
    \includegraphics[width=\linewidth]{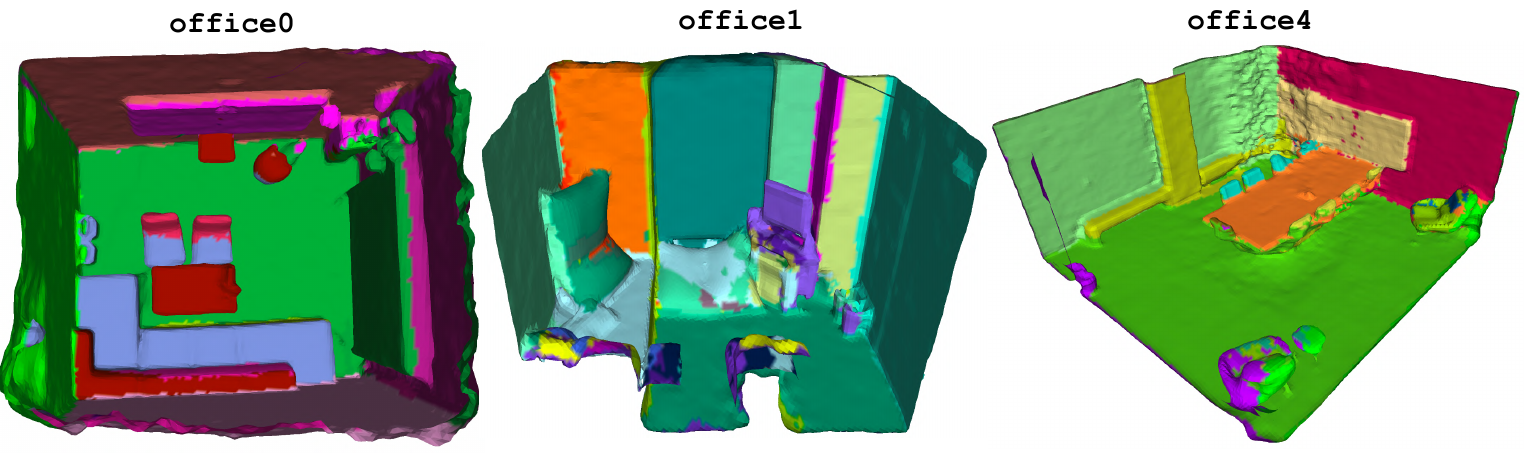}
    \caption{Replica}
    \label{fig:replica}
  \end{subfigure}
  \hfill
  \begin{subfigure}[b]{0.8\linewidth}
    \includegraphics[width=\linewidth]{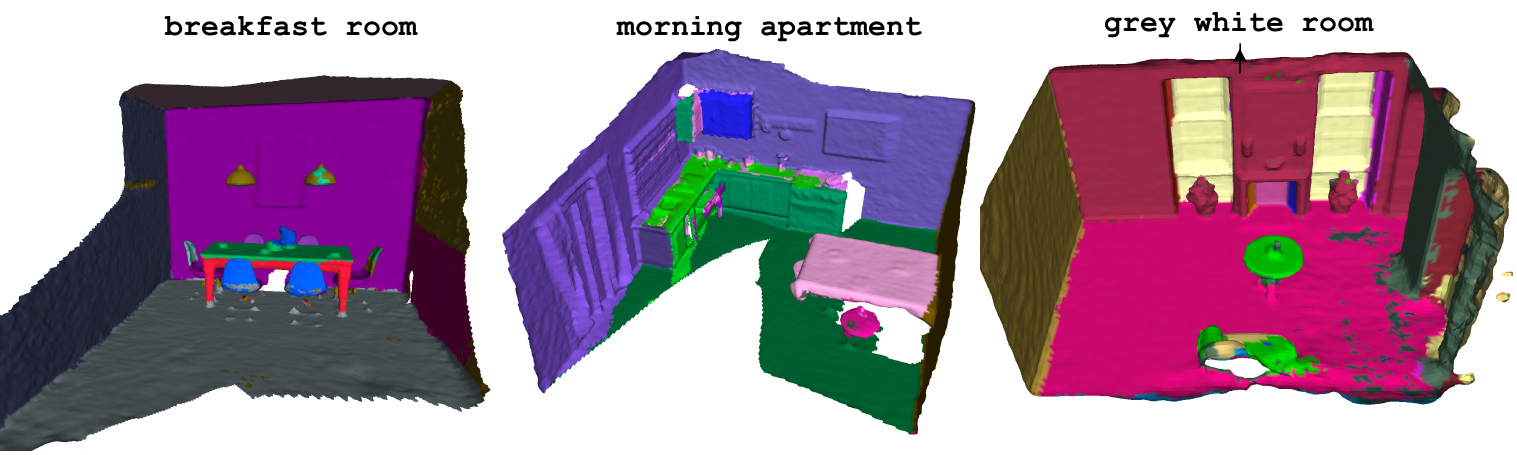}
    \caption{Synthetic}
    \label{fig:synthetic}
  \end{subfigure}
  \caption{Results by \titlePrefix{} for (a) Replica dataset, and (b) Synthetic dataset.}
  \label{fig:other_data}
\end{figure}


From the above quantitative results, we can observe that PlanarRecon achieves the best performance among all the baselines. To further validate the advantages of our method, we show more \textbf{qualitative} comparisons between  \titlePrefix{} with PlanarRecon in Fig.~\ref{fig:qual_2}. Both of our methods (\titlePrefix{}-SS and \titlePrefix{}-S) maintain high-quality performance across diverse indoor rooms.

\subsection{Ablation Studies\protect\footnotemark[2]}
\footnotetext[2]{For the purpose of ablation, we randomly select 10 scenes from the validation set. The results of all \textit{quantitative} experiments through this section are based on the selected scenes.}
\label{sect:ablation}
\noindent\textbf{Replica and Synthetic.} We show qualitative results of our model on Replica and Synthetic datasets in Fig.~\ref{fig:other_data}. Our model can generate excellent plane reconstructions without any annotations (pose/2D planes/3D planes) in an online manner. Note that there is no ground truth and none of the baselines reported results for those datasets. Therefore, we are only able to show the results from \titlePrefix{}-SS. More results by our model on those datasets are listed in the supplementary material.\\
\begin{figure} [t]
  \begin{center}
    \includegraphics[width=0.8\columnwidth]{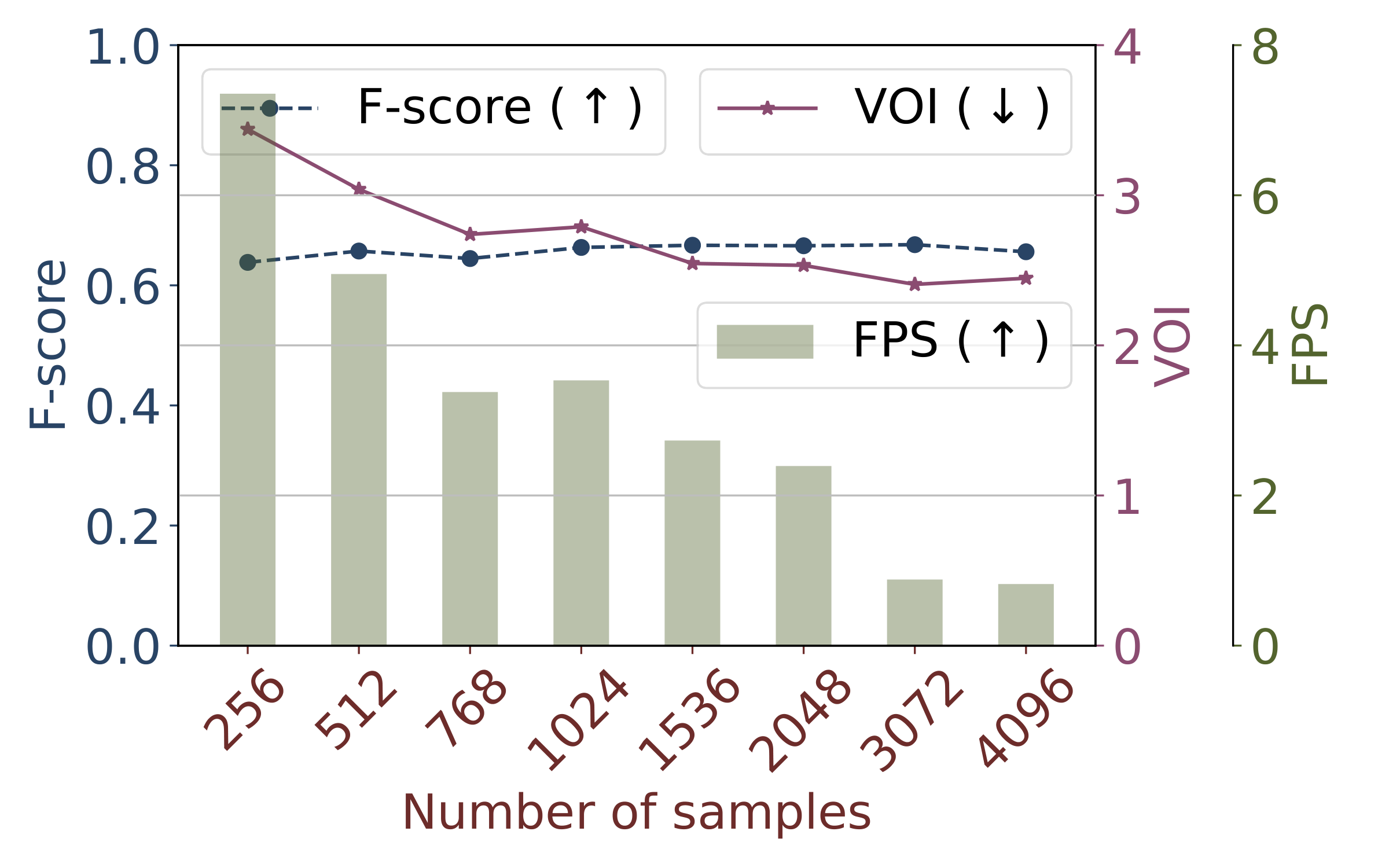}
  \end{center}
  \vspace{-15pt}
  \caption{\small Ablation for the number of samples used in \titlePrefix{}. \vspace{-15pt}
  } 
  \label{fig:ablation_number}
\end{figure}
\textbf{How many samples are used?} The number of samples used in \titlePrefix{} is very important because they are used for all included learning modules (pose/SDF/color/plane) in the proposed framework. It is also closely related to the computational speed. To achieve the best tradeoff, we have conducted thorough experiments. The detailed comparisons are presented in Fig.~\ref{fig:ablation_number}, where we report the geometry quality with F-score; segmentation quality with VOI; and the speed with Frames Per Second (FPS). In our work, 
$768$ samples are used.
\\ 
\noindent\textbf{Plane similarity measure.} To validate the usefulness of Eq.~(\ref{eq:plane_dist_2}) and show the disadvantage of the Eq.~(\ref{eq:plane_dist_1}), we quantitatively compare different plane similarity measures in Table~\ref{tab:similarity_measure}, from where we can see that using Eq.~(\ref{eq:plane_dist_2}) achieves the best performance.\\
\begin{table}[t]
\centering
\resizebox{\columnwidth}{!}
{ %
\begin{threeparttable}
\caption{Ablation studies for similarity measurement. 
$\diamond$: Directly applying Euclidean distance to raw plane parameters.
} 
\small
\label{tab:similarity_measure}
\begin{tabular}{l|ccc}
\hline \hline 
 Method & VOI~$\downarrow$ & RI~$\uparrow$ & SC~$\uparrow$\\
\hline
Raw plane param.\tnote{$\diamond$} & 3.368 & 0.821 & 0.132\\
Corrected plane param. (Eq. (\ref{eq:plane_dist_1})) & \textbf{\color{teal}3.017} & \textbf{\color{teal}0.829} & \textbf{\color{teal}0.200}\\
Points-to-plane dist. (Eq. (\ref{eq:plane_dist_2})) & \textbf{\color{purple}2.833} & \textbf{\color{purple}0.857} & \textbf{\color{purple}0.319} \\
\hline \hline
\end{tabular}
\end{threeparttable}
}
\end{table}
\begin{figure}[t]
  \centering
   \includegraphics[width=0.9\linewidth]{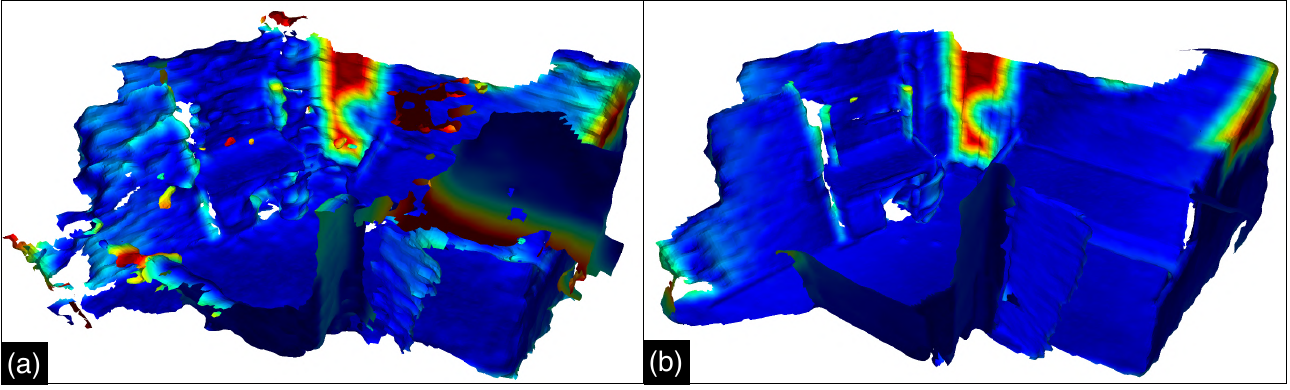} 
   \caption{\small Error map with (a) allowing gradients backpropagation and (b) blocking gradients backpropagation. Red color means a high error and blue color means a low error. Note that the dark red region appears in (a) and (b) because the ground truth fails to capture the window area. \vspace{-15pt}
   } 
   \label{fig:ablation_grad}
\end{figure}
\textbf{Thershold for similarity measure.} After the computation of the similarity measure, we need a threshold to determine whether the two planes belong to one instance. If the threshold is too small, there will be too much noise. If the threshold is too large, parallel planes might be treated as one instance. We use the threshold of $0.1$ (See Table~\ref{tab:similarity_threshold}).\\
\textbf{Coefficient for EMA.} During the maintenance of the global memory bank, we use an EMA to update the plane parameters in the bank. The selection of the coefficient in EMA can also affect the final performance a lot. We take the value of $\psi$ as $0.999$. Please see a quantitative comparison in Table~\ref{tab:ema}.\\
\textbf{Gradient Backpropagation.} In \titlePrefix{} model architecture (Fig.~\ref{fig:overall_framework}), we stop backpropagating the gradients from the plane branch to the SDF branch during training. This is necessary because the gradients from plane rendering loss can disturb the training of the SDF MLP, weakening the reconstruction quality. We show the qualitative comparison using error maps in Fig.~\ref{fig:ablation_grad}.

\section{Conclusion}
In this paper, we propose a novel plane detection model, \titlePrefix{}. This framework introduces a unique methodology that combines plane segmentation rendering, an efficient plane fitting module, and an innovative memory bank for 3D planar detection and global tracking. These contributions enable \titlePrefix{} to learn effectively from monocular RGB and depth sequences. Demonstrated through extensive testing, its ability to outperform existing methods marks a significant advancement in plane detection techniques. \titlePrefix{} not only challenges existing paradigms but also sets a new standard in the field, highlighting its potential for diverse real-world applications.

\bibliographystyle{unsrt}
\bibliography{ref}

\end{document}